\documentclass[runningheads]{llncs}
\usepackage{graphicx}
\usepackage{caption}
\usepackage{subfigure}
\usepackage{makecell}
\usepackage{multirow}

\usepackage{tikz}
\usepackage{comment}
\usepackage{amsmath,amssymb} 
\usepackage{color}

\usepackage[accsupp]{axessibility}  

\begin{document}
\pagestyle{headings}
\mainmatter
\def\ECCVSubNumber{3834}  

\title{Switch-BERT: Learning to Model Multimodal Interactions by Switching Attention and Input} 

\titlerunning{Switch-BERT}
%
\author{Qingpei Guo\inst{1} \index{Qingpei, Guo} \and
Kaisheng Yao\inst{2} \index{Kaisheng, Yao} \and
Wei Chu\inst{1}}
\authorrunning{Q. Guo et al.}
%
\institute{Ant Group \email{\{qingpei.gqp, weichu.cw\}@antgroup.com} \and
Amazon AWS AI \email{kaishey@amazon.com}}
\maketitle

\begin{abstract}
The ability to model intra-modal and inter-modal interactions is fundamental in multimodal machine learning. The current state-of-the-art models usually adopt deep learning models with fixed structures. They can achieve exceptional performances on specific tasks, but face a particularly challenging problem of modality mismatch because of diversity of input modalities and their fixed structures. In this paper, we present \textbf{Switch-BERT} for joint vision and language representation learning to address this problem. Switch-BERT extends BERT architecture by introducing learnable layer-wise and cross-layer interactions. It learns to optimize attention from a set of attention modes representing these interactions. One specific property of the model is that it learns to attend outputs from various depths, therefore mitigates the modality mismatch problem. We present extensive experiments on visual question answering, image-text retrieval and referring expression comprehension experiments. Results confirm that, whereas alternative architectures including ViLBERT and UNITER may excel in particular tasks, Switch-BERT can consistently achieve better or comparable performances than the current state-of-the-art models in these tasks. Ablation studies indicate that the proposed model achieves superior performances due to its ability in learning task-specific multimodal interactions. 
\end{abstract}

\section{Introduction}
\label{sec:introduction}

\begin{figure}[htbp]

\subfigure[Illustration of modality misalignment in VilBERT(fixed structure).]{ 
\begin{minipage}[t]{0.48\textwidth}
\centering  
\includegraphics[width=6cm, height=0.7\linewidth]{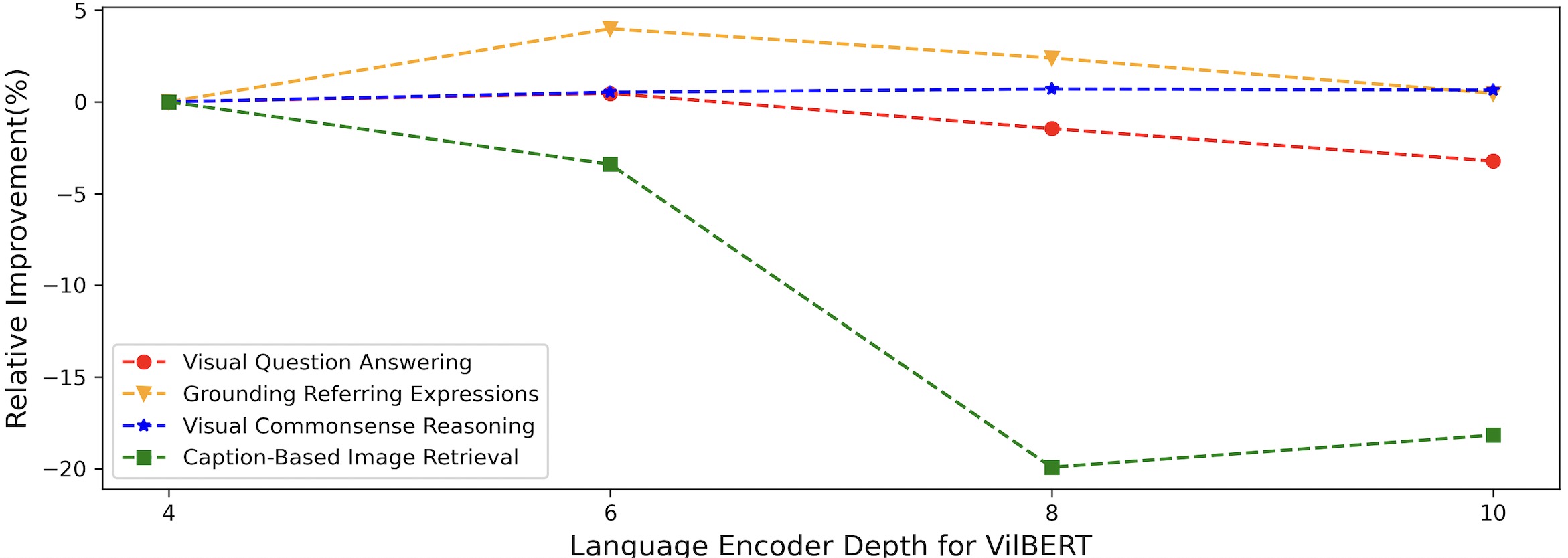}
\end{minipage}}
\subfigure[Comparison between popular multimodal architectures with Switch-BERT.]{
\begin{minipage}[t]{0.48\textwidth}
\centering
\includegraphics[width=6cm]{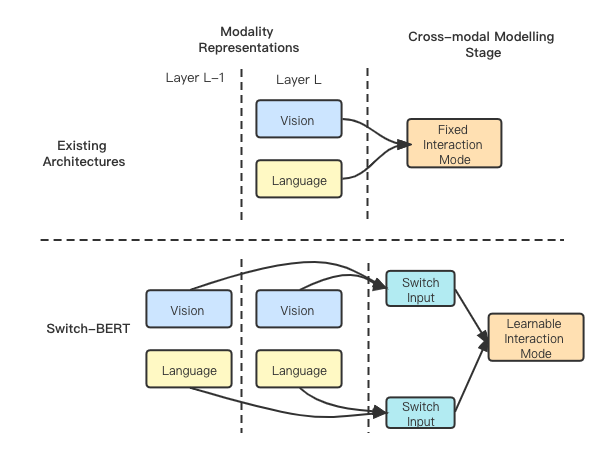}
\end{minipage}}

\caption{ (a) The text and visual encoder are separate before their interactions in VilBERT~\cite{lu2019vilbert}. By varying the depth of text encoder from 4 to 10, the accuracy of VilBERT is changed relative to that from depth 4. The optimal relative improvements in accuracy are different in the four tasks. For VilBERT, the misalignment can degrade performances by approximately 20\% relatively. (b) In contrast to fixed structures, Switch-BERT learns to attend outputs from various depth and has learnable layer-wise and cross-layer interactions. }   
\label{fig:1}  
\end{figure}

The current state-of-the-art approaches for multimodal machine learning~\cite{chen2020uniter,li2020unicoder,li2019visualbert,lin2020interbert,lu2019vilbert,su2019vl,tan2019lxmert} are based on the BERT encoders~\cite{devlin2018bert} that use the Transformer architecture~\cite{vaswani2017attention}. These BERT-based models follow two design paradigms for intra-modal and inter-modal interactions. The first paradigm utilizes a single-stream BERT encoder to jointly encode representations from these modalities, such as those from vision and language~\cite{chen2020uniter,li2020unicoder,li2019visualbert,lin2020interbert,su2019vl}. In this case, intra-modal interactions and the implicit association between modalities are jointly modeled with the multi-head attention mechanism~\cite{vaswani2017attention}. The second paradigm learns modal-specific representations through different BERT encoders, for instance using dual-stream BERT encoders on vision and language~\cite{lu2019vilbert,tan2019lxmert}. These methods achieve inter-modal interactions via specially designed structures such as cross-attention sub-layers~\cite{lu2019vilbert,kazemzadeh2014referitgame,yu2018mattnet}.

However, misalignment between modal semantics is a challenging problem for these methods. For example, the visual modality observation often is based on region-level semantic feature from detection models such as Faster R-CNN~\cite{ren2015faster}, whereas the text modality observation can be simply raw tokens or sub-word tokens such as word-pieces~\cite{wu2016wordpiece}. For single-stream models, these visual features with high-level semantics and text input with low-level semantics are both fed to the BERT encoder simultaneously. Given that these observations are not at the same semantic level, using a common encoding process for different modalities seems to be contradictory. The dual-stream models can ease the misalignment problem with distinct encoding process for each modality. 
However, the interaction between modalities of dual-stream models is restricted to specific layers that can be inflexible. 

Fig.~\ref{fig:1}(a) illustrates this problem by tuning the depth of a BERT-based language encoder before interaction with the visual stream in ViLBERT~\cite{lu2019vilbert} on a set of tasks. Though with deeper encoder that usually extracts higher level semantics~\cite{2019Open,tenney-etal-2019-bert}, performances don't reveal monotonous trend with the depth. This indicates that misalignment between modal semantics poses challenges to optimal multimodal performances. Another observation in Fig.~\ref{fig:1}(a) is that the optimal depths are different for these tasks, indicating that a fixed architecture is hardly optimal for every task. This suggests necessity for more flexible architectures. The modality misalignment problem is however not well studied. 

In this paper, we propose Switch-BERT to alleviate the modality misalignment problem. As illustrated in Fig.~\ref{fig:1}(b), Switch-BERT extends the recently developed multimodal methods but has sample-specific interactions among modalities, instead of fixed architectures adopted in the previous approaches for every sample. Specifically, it introduces two modules, respectively for layer-wise switch operation in Switch-Attention Block (SAB) and cross-layer switch operation in Switch Input Block (SIB). 
The SAB module learns to attend to, given a sample, particular modality and choose from a set of predefined operations for interactions among modalities. 
The SIB module introduces sample-specific modeling of cross-layer modal representations and learns to switch inputs among representations at various depths. 

We pre-train Switch-BERT on Conceptual Captions~\cite{sharma2018conceptual} to learn task independent visual and text grounding. Proxy pre-training tasks include masked language modeling with visual clues (MLM), masked region classification with KL-divergence (MRC-KL)~\cite{chen2020uniter} and Image-Text Matching (ITM). 
We evaluate Switch-BERT on three downstream tasks including visual question answering, cross-modal retrieval and referring expression comprehension, and perform experiments on  VQAv2~\cite{goyal2017making}, Flick30k~\cite{plummer2015flickr30k} and RefCOCO+~\cite{kazemzadeh2014referitgame} datasets. Experimental results show Switch-BERT can learn better multimodal representations, compared with previous single- and dual-stream models.
We conduct ablation studies and show that Switch-BERT can learn task-specific multimodal interactions end-to-end, including layer-wise interaction selection and cross-layer input selection. This task-specificity is an advantage over other methods with fixed architectures.

\section{Methodology}

\subsection{Preliminaries}
\subsubsection{Language BERT Encoder.} BERT~\cite{devlin2018bert} was originally proposed for natural language processing tasks to learn semantic representations for each input token via a stack of transformers~\cite{vaswani2017attention}. A BERT encoder consists of $L$ transformer layers, in which representation $X_l$ at $l$-th layer is obtained from the representation ${X_{l-1}}$ in its lower layer as follows:

\begin{eqnarray}
  {X}_{l}& = &LN(\bar{X}_{l} + GeLU(\bar{X}_{l}W_{1})W_{2}), \label{eqn:FFblock} \\
\bar{X}_{l} & = & LN(\hat{X}_{l} + X_{l-1}), \label{eqn:LayerNorm}  \\
 \hat{X}_{l} & = & MHA(Q_l, K_l, V_l) \label{eqn:MultiHeadAttention}
 \end{eqnarray}
\noindent where $\textit{MHA}(\cdot)$ implements the multi-head attention mechanism~\cite{vaswani2017attention}, with query, key, and value at layer $l$ each computed as $Q_l = X_{l-1}W^{Q}$, $K_l = X_{l-1}W^{K}$, and $V_l = X_{l-1}W^{V}$.  $\textit{LN}$ is layer normalization~\cite{ba2016layer}, $\textit{GeLU}$~\cite{hendrycks2016gelu} is the activation function of feed forward block. $W^Q, W^K \in R^{d \times d^{q}}$,  $W^V \in R^{d \times d^i}$,  and $W_{1}, W_{2}^{\top} \in R^{d \times d^{f}}$ are learnable matrices. The multi-head attention block and feed forward block form a transformer layer.

\subsubsection{Multimodal BERT Encoder.} Multimodal BERT~\cite{kiela2019supervised,lu2019vilbert} extends the language BERT with multimodal input vector sequences. For instance, for tasks that consist of image and text, the model assigns two types of inputs: image can be a sequence of vectors as $X^i = [\textit{IMG},  i_{1}, \cdots , i_{N_i-1}] \in R^{N_i \times d_i}$ and text can be $X^t = [\textit{CLS}, w_{1}, \cdots, w_{N_{t}-2}, \textit{SEP}] \in R^{N_{t} \times d_{t}}$, where \textit{IMG}, \textit{CLS} and \textit{SEP} are embeddings of special markers. Usually, we have $d_i=d_{t}=d$. 
Typical approaches include UNITER~\cite{chen2020uniter}, in which $X^i$ and $X^t$ are concatenated, forming a single stream of input $X_0 = [X^i X^t]$ to compute query, key and value matrices. In contrast, ViLBERT~\cite{lu2019vilbert} computes query from one modality but key and value from other modality, and vice versa, forming dual steams of computations.

\subsection{Generalizing BERT Encoder}
\label{sec:generalization}

\begin{table}[b]
\begin{center}
	\caption{Multimodal interaction mode spaces} 
	\label{tab:interactionMode}
	\begin{tabular}{lll}
	\noalign{\smallskip}
	\hline
	\makecell[c]{Interaction Mode} & &\makecell[c]{Attention Mechanisms} \\
	\noalign{\smallskip}
	\hline
	$M_0$: Self-Self Attention & & $X$ \& $\urcorner X$: Self-Attention \\
	\hline
	$M_1$: Self-Cross Attention & &$X$: Self-Attention, $\urcorner X$: Cross-Attention \\
	\hline
    $M_2$: Cross-Self Attention & & $X$: Cross-Attention, $\urcorner X$: Self-Attention \\
	\hline
    $M_3$: Joint-Attention & &$X$ \& $\urcorner X$: Joint-Attention \\
    \hline
		\end{tabular}
	\end{center}
	\end{table}

We would like to generalize the encoder in Eqs. (\ref{eqn:FFblock}-\ref{eqn:MultiHeadAttention}) beyond the multimodal architectures described above. To this end, we first use $X \in \{X^i, X^{t}\}$ to denote either the image modality observation $X^i$ or text modality observation $X^{t}$. We use $\urcorner X$ to denote complementary of $X$; e.g., $\urcorner X = X^i$ if $X = X^t$. Notice that $X^i$ and $X^t$ are for purpose of notations, and can be generalized beyond image and text modalities. 

We further generalize the multi-head attention mechanism in Eq. (\ref{eqn:MultiHeadAttention}) beyond linear projections on input $X_{l-1}$, in which query, key, and value are obtained via certain transformations. Formally, we rewrite Eq. (\ref{eqn:MultiHeadAttention}) as follows: 
\begin{equation}
 \hat{X}_{l}=MHA(q(X_{input}),k(X_{context}),v(X_{context})), \label{eqn:MultiHeadAttentionRewrite}
\end{equation}
\noindent where $q(\cdot)$, $k(\cdot)$ and $v(\cdot)$ extract query, key and value representations, respectively. Notice that key and value operations share the input observation $X_{context}$, whereas query $q(\cdot)$ operates on $X_{input}$. 

Eq. ($\ref{eqn:MultiHeadAttentionRewrite}$)  enables us to relate the previously proposed multimodal approaches. For dual-stream models~\cite{lu2019vilbert,tan2019lxmert}, intra-modal and inter-modal interactions are independently modeled explicitly. Using Eq. ($\ref{eqn:MultiHeadAttentionRewrite}$), Self-Attention for intra-modal interaction is modeled with $ X_{input} = X_{l-1}$ and $X_{context} = X_{l-1}$. Cross-Attention for inter-modal interaction can be achieved using $X_{input} = X_{l-1}$ and $X_{context} = \urcorner X_{l-1}$. For single-stream models~\cite{li2020unicoder,chen2020uniter,li2019visualbert}, intra-modal and inter-modal interactions are implicitly modeled with Joint-Attention using $X_{input} = X_{l-1}$ and $X_{context} = [X_{l-1},\urcorner X_{l-1}]$, the latter is obtained via concatenation and enables attention to the whole multimodal context.

Among the above described attention mechanisms, Joint-Attention uses whole multimodal context, therefore has potential of representation of both Self-Attention and Cross-Attention. However, its multimodal context can face potential semantic misalignment between modalities, as described in Sec.~\ref{sec:introduction}. On the other hand, Self-Attention and Cross-Attention restrict the modal context to attend, easing semantic misalignment of modalities, but leads to limited representation to particular modality.

We therefore design a more complete space of multimodal interactions. Table~\ref{tab:interactionMode} lists four interaction modes between $X$ and its complementary $\urcorner X$.  Self-Self Attention invokes self-attention on each modality. Self-Cross Attention has $X$ use Self-Attention and $\urcorner X$ use Cross-Attention, and vice versa for Cross-Self Attention. Joint-Attention has both $X$ and $\urcorner X$ conduct their own Joint Attention operations. Those attention operators share the same attention weights in our setting and can be implemented with different layer-wise attention masks.

\subsection{Switch Attention and Input Block}
\label{sec:saib}

\begin{figure}[tbp]
\centering 
\subfigure[Illustration of Switch-BERT layer and Switch-Attention Block.]{   
\begin{minipage}[t]{0.48\textwidth}
\centering  
\includegraphics[width=6cm]{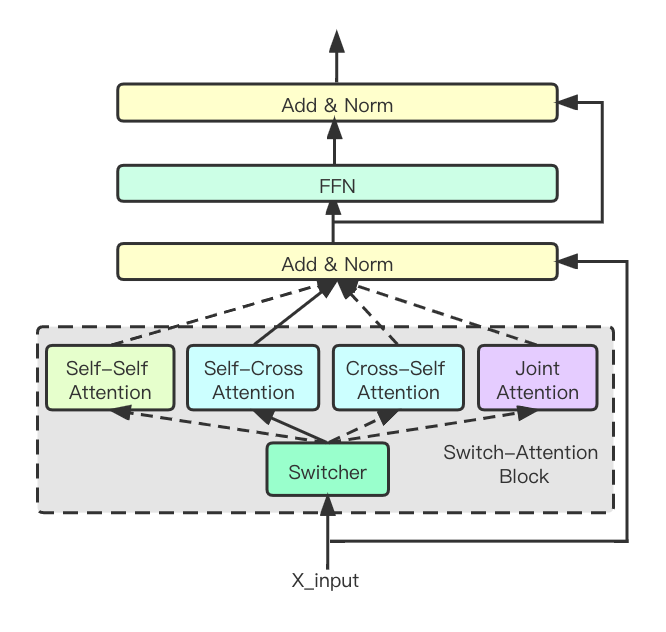} 
\end{minipage}}
\subfigure[Illustration of Switch-Input Block.]{ 
\begin{minipage}[t]{0.48\textwidth}
\centering   
\includegraphics[width=6cm, height=1\linewidth]{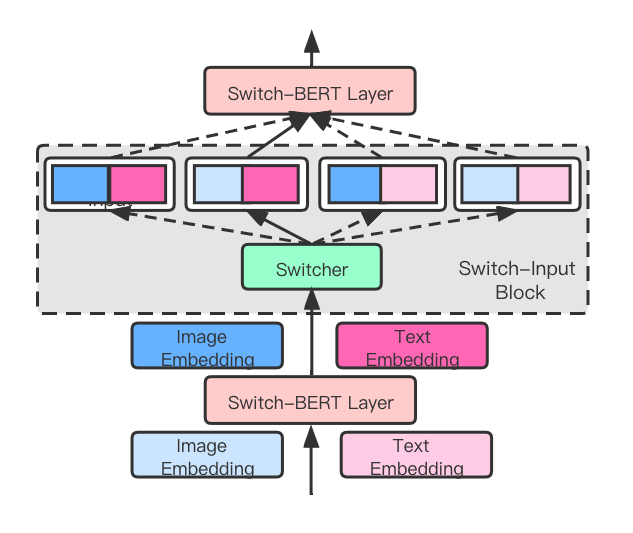}
\end{minipage}
}
\caption{ (a) The Switch-BERT layer extends the Multi-head Joint Attention block in a normal transformer encoder layer with our proposed Switch-Attention Block. (b) The Switch-Input Block brings in modality representations from current and previous layers for its successive Switch-BERT layer. (best viewed in color)}    
\label{fig:2}   
\end{figure}

\subsubsection{Switch-Attention Block.} 
Unlike conventional multimodal models that limit the modality interaction between specific layers, we employ the Switch-Attention Block (SAB) to achieve learning layer-wise multimodal interaction in an end-to-end manner. As illustrated in Fig.~\ref{fig:2} (a), SAB depends on an attention switcher module to search for an appropriate mode from the multimodal interactions described in Table~\ref{tab:interactionMode}. The search space can be formally defined as a set of operations $\{M_n\}_{n=1}^{N_{a}}$, where $N_{a}$ indicates the number of interaction modes. 

We describe in the following a switcher method to search for proper interaction. Given a holistic representation of image and text as $X_{l}^{\textit{i}}$ and $X_{l}^{\textit{t}}$,  we apply average pooling over the image and text tokens to obtain global features of each modality:
\begin{equation}
z_{l}^{i} = AvgPool(X_{l}^{\textit{i}}), z_{l}^{t} = AvgPool(X_{l}^{\textit{t}}))).
\end{equation}
Then we define the modality “alignment degree" of the $l$-th layer as $d_{l} = z_{l}^{i} \odot z_{l}^{t}$, and apply a trainable MLP, $f_{MLP}$ with Softmax activation to obtain the probability of the interaction modes $\pi$:
\begin{equation}
\pi = Softmax(f_{MLP}(d_{l})).
\label{eqn:Softmax}
\end{equation}

We use Gumbel-Softmax  reparameterization \cite{Gumbel-Softmax} to sample a particular interaction based on the above probability, in which probability of interaction $M_n$ is
\begin{equation}
 p(M_n) = \frac{\exp{((\log{(\pi_n)}+g_n)}/\tau)}{\sum_{j=1}^{N_{a}}\exp\left((\log(\pi_{j})+g_{j})/\tau\right)},
\label{eqn:Gumbel-Softmax}
\end{equation}
\noindent where $g_n$ is sampled Gumbel noise, computed as $g_n = -\log(-\log(u_n))$, with $u_n$ sampled from uniform distribution of $\textit{Uniform}(0,1)$. $\tau$ is the smooth parameter for Gumbel-Softmax distribution. 

Given $X_{input}=X_{l}^{\textit{i}} \cup X_{l}^{\textit{t}}$, SAB performs “soft weighting" or “hard selection" of interaction modes by:
\begin{equation}
\begin{aligned}
y_{soft} &= \sum_{i\in N_{a}}{p(M_{i})M_{i}(X_{input})} \\
y_{hard} &= M_{n^{\star}}(X_{input}), n^{\star}=\underset{n}{\mathrm{argmax}}\{p(M_n)\}
\label{eqn:SwitchAttentionOutput}
\end{aligned}
\end{equation}
. 

For training, we start at a high temperature in Eq.\ref{eqn:Gumbel-Softmax} for small gradient variance, then anneal to a small but non-zero temperature to make the output distribution $p(M)$ approximate one-hot. We adopt “soft weighting" of attention modes during training and “hard selection" for inference.

\subsubsection{Switch-Input Block.} 
To ease semantic misalignment between modalities, we propose Switch-Input Block (SIB) to bring in cross-layer modal representation. 
SIB enables Switch-BERT layer, illustrated in Fig.~\ref{fig:2} (a), to take input either from the output of its lower layer or from the residual connection in the lower layer, which connects to the output from the layer further below. Concretely, for $l$-th layer with $l \ge 2$, its input is in a set of $\{X_{l-1} \urcorner X_{l-1}\}$ $\cup$ $\{X_{l-1} \urcorner X_{l-2}$, $X_{l-2} \urcorner X_{l-1}$, $X_{l-2} \urcorner X_{l-2}\}$. We then apply switch operation on the set and obtain an element from the set as input $X_{input}$ to layer $l$.
The switcher algorithm follows Eq. \ref{eqn:Softmax}, Eq.\ref{eqn:Gumbel-Softmax} and Eq. \ref{eqn:SwitchAttentionOutput} but is trained specifically for SIB. Fig.~\ref{fig:2} (b) illustrates the Switch-Input Block. 

\subsection{The Switch-BERT Model}

 The Switch-BERT model's components are described in details below. Further details are in the supplementary material.

\subsubsection{Visual and Text Embedding.} Following \cite{lu2019vilbert}, images are represented with detected objects. We extract the bounding box and visual feature of each object from the widely used Faster-RCNN~\cite{ren2015faster} detector trained on Visual Genome~\cite{Visualgenome}.  We also add a type field (VisualType/TokenType) to distinguish visual and text input. The region feature, position and type field are fed into a visual embedding layer to obtain the visual embedding for Switch-Encoder. A special IMG token representing the entire image segment is also inserted at the beginning of the visual sequence. The text embedding is generated following BERT~\cite{devlin2018bert}, in which we tokenize the input sentence and keep orders of tokens as their position ids. The token, position and type field are fed into a text embedding layer to perform embedding lookup.     

\subsubsection{Switch-Encoder.} Given the pair of visual and text embedding, the Switch-Encoder learns to model layer-wise multimodal interactions. The Switch-Encoder consists of a stack of Switch-BERT layers, with Switch-Input Block inserted between consecutive Switch-BERT layers. Switch-BERT layer in Fig.~\ref{fig:2} (a) generally follows the architecture of the Transformer encoder layer~\cite{vaswani2017attention}, but distinguishes it with the adaptive multimodal attention mechanism using Switch-Attention Block. It takes the entire representations from visual and text embedding, but selects sample-specific interactions of these representations. The Switch-Input Block routes the modality input for the following Switch-BERT layer to help alleviate semantic misalignment. The rest of the Switch-Encoder proceeds similarly as that in BERT encoder, resulting in a multimodal feature as its output.

\subsubsection{Pretraining Tasks.} Task-agnostic multimodal pre-training can help learn associations between modalities. Like previous work~\cite{lu2019vilbert,li2019visualbert,li2020oscar,miech2020end,sun2019videobert,zhou2020unified,lin2020interbert}, we first pre-train Switch-BERT on proxy tasks and then adapt it to downstream tasks through finetuning. Three proxy tasks are used for pre-training. (1) Masked language modeling with visual clues (MLM). This task follows the MLM objective in BERT~\cite{devlin2018bert} but with the above described contextualized multimodal input. In this task, word tokens are randomly masked but with their positions preserved. The model needs to predict the token from the left visual and textual context. (2) Masked region classification with KL-divergence (MRC-KL)~\cite{chen2020uniter}. Similar to MLM, this task masks approximately 15\% of the region features. MRC-KL then trains the model to predict the class distribution from the object detector for the region, rather than reconstructing the feature of masked regions.  (3) Image-Text matching (ITM). Given paired image-and-text as positives, their negative pairs are generated by randomly replacing texts in the positive pairs with unrelated ones. 
The ITM task is for the model to distinguish positive pairs from negatives.

\section{Experiments}

\subsection{Datasets and Downstream Tasks}
 We evaluate Switch-BERT on different types of downstream tasks including image-text retrieval, referring expressions and vocab-based VQA. Their statistics are shown in Table~\ref{tab:Datasets}.

\begin{table}[t]
	\caption{Statistics of Datasets for the Downstream Tasks}
	\label{tab:Datasets}
	\begin{center}
	\begin{tabular}{ccccc}
	\noalign{\smallskip}
	\hline
	\textbf{Dataset}&\textbf{Tasks}&\textbf{Train} & \textbf{Test} &\textbf{Metric}\\
	\noalign{\smallskip}
	\hline 
	Flick30k & Image-Text Retrieval& 29k & 1k & Recall@k \\
	\hline 
	RefCOCO+ & Referring Expression & 120k & 10.6k & Accuracy \\
	\hline 
	VQAv2 & Visual Question Answering& 657k & 107.3k & VQA-score \\
	\hline 
	\end{tabular}
	\end{center}
	\end{table}

 \subsubsection{Image-Text Retrieval.}  Given images or captions, the image-text retrieval task requires the model to perform cross-modal retrieval. We conduct experiments on Flick30k~\cite{plummer2015flickr30k} dataset, which has images paired with five captions. Following~\cite{lu2019vilbert}, we train models on Flick30k in a 4-way multiple-choice setting. For each image-text pair, three negatives are generated by replacing the caption with a random one and replacing the image with a random and a hard one. The model outputs similarity scores of these four image-text pairs as the ITM task. Once softmax is computed on the similarity scores, cross-entropy loss is applied to learn the models. We report Recall@1.

 \subsubsection{Referring Expressions Comprehension.} This task focuses on localizing objects queried by a natural language expression. For the RefCOCO+~\cite{kazemzadeh2014referitgame} dataset, we take the bounding boxes detected by \cite{yu2018mattnet} and select the top 36 regions with the highest class scores. Following the conventions in~\cite{su2019vl,lu2019vilbert}, a simple fully-connected layer is added on top to regress the matching degree, defined as the IOU with the ground truth box, with the referring expression for each input region. We train the model with binary cross-entropy loss. To evaluate, regions with matching degree above threshold of 0.5 are considered correct. We apply the accuracy score as the evaluation metric.
 
\subsubsection{Visual Question Answering.} Given questions about an image, this task expects the model to give correct answers. Following \cite{anderson2018bottom}, we consider the VQA~\cite{goyal2017making} task a multi-label classification problem on a closed answer pool and generate the target soft-label based on its relevance to ten human answer responses. We add two fully-connected layers to map the multimodal representation, which is the element-wise product fusion of image and text representation, to the answers' space and apply binary cross-entropy loss for training. Following the same protocol with SOTA baselines, we train models on train-val split and report VQA-score~\cite{antol2015vqa} on the test-dev split.

\setcounter{footnote}{0}
\subsection{Controlled Settings}
Shown in \cite{lu2019vilbert,tan2019lxmert,chen2020uniter,qi2020imagebert},  the quality and volume of the pre-training data significantly impact the performance of multimodal BERTs. This explain most of the claimed performance differences in downstream tasks~\cite{bugliarello2021multimodal}. In this paper, we focus our discussion on the independent contribution of architecture design. To exclude performance influences other than architectures and enable fair comparison under limited resources, we adopt the controlled settings introduced by \cite{bugliarello2021multimodal}. Specifically, we pre-train multimodal BERTs on the same subset of 2.7M image-text pairs of Conceptual Captions~\cite{sharma2018conceptual} for 10 epochs and employ the same proxy tasks as our Switch-BERT model.  We use the VOLTA~\footnote{https://github.com/e-bug/volta}  implementation for all state-of-the-art models for comparison in our experiments, and train these multimodal BERTs with a fixed set of hyperparameters, such as encoder dimensions, methods for modality fusion, number of MLP layers in the finetune head, to exclude possible confounds that may interfere with a fair comparison of these architectures. Models with the best validation set  performance are chosen for downstream tasks evaluation~\footnote{We train with three different random seeds and report their average performances}.
Due to space constraints, more implementation details as well as hyper-parameter settings are split into the supplementary materials.

\subsection{Main Results}

\begin{table}[t]
\setlength{\abovecaptionskip}{-0.5cm}
	\caption{Results on downstream tasks. We adopt the re-implementation from the VOLTA\cite{bugliarello2021multimodal} framework for baseline models. All models perform the same controlled settings and “*” denotes models without pre-training on Conceptual Captions\cite{sharma2018conceptual}. We report std of Switch-BERT as well as baseline models on three runs with different random seeds.}
	\label{tab:main_results}
	\begin{center}
	\resizebox{\textwidth}{22mm}
 {
	\begin{tabular}{ccccccc}
	\noalign{\smallskip}
	\hline
	\multicolumn{2}{c}{\multirow{2}{*}{\textbf{Models}}} &
	\multirow{2}{*}{Params} &
	\multirow{2}{*}{VQAv2} & \multicolumn{2}{c}{Flick30K-Retrieval}& \multirow{2}{*}{RefCOCO+}\\
	& & & &  Image Retrieval & Text Retrieval  & \\
	\noalign{\smallskip}
	\hline
	\multirow{3}{*}{\shortstack{Single-stream \\ (Fixed) }} & UNITER\cite{chen2020uniter} &  114.9M & 68.8 $\pm$ 0.4 & 60.9 $\pm$ 0.7 & 76.4 $\pm$ 1.3 &  71.9 $\pm$ 0.67\\
 
	& VL-BERT\cite{su2019vl} & 116.1M &  68.3 $\pm$ 0.31 & 57.9 $\pm$ 1.1 & 70.9 $\pm$ 1.7 & 71.1 $\pm$ 0.23 \\

	& VisualBERT\cite{li2019visualbert} & 114.9M &  68.9 $\pm$ 0.27 & 61.1 $\pm$ 1.2 & 75.5 $\pm$ 1.8 & 69.7 $\pm$ 0.31 \\
	\hline
	\multirow{2}{*}{\shortstack{Dual-stream \\ (Fixed)}} & 	LXMERT\cite{tan2019lxmert} & 211.4M &  67.1 $\pm$ 0.34 & 58.6 $\pm$ 1.4 & 74.9 $\pm$ 2.7 & 69.8 $\pm$ 0.46 \\

	& VilBERT\cite{lu2019vilbert} & 242.1M & 68.7 $\pm$ 0.82 & 59.8 $\pm$ 0.8 & \textbf{78.3} $\pm$ 1.6 &  70.8 $\pm$ 0.58 \\
 
	\hline
  
	\multirow{2}{*}{Dynamic} & Switch-BERT$^{*}$ &  \multirow{2}{*}{130.6M} & 66.7  $\pm$ 0.97 & 38.2  $\pm$ 1.7 & 57.3  $\pm$ 2.3 &  68.9  $\pm$ 0.82   \\
	& Switch-BERT &  & \textbf{70.7} $\pm$ 0.62 & \textbf{62.2} $\pm$ 0.9 & 78.2 $\pm$ 1.6 &  \textbf{72.8}  $\pm$ 0.45  \\  
	\hline
	\end{tabular}
 }
    \end{center}
	\end{table}

We compare the proposed Switch-BERT against existing multimodal architectures of both single and dual-stream on three widely-used benchmark datasets. Baselines for comparison include the state-of-the-art multimodal architectures of ViLBERT~\cite{lu2019vilbert}, UNITER~\cite{chen2020uniter}, VisualBERT~\cite{li2019visualbert}, VL-BERT~\cite{su2019vl} and LXMERT~\cite{tan2019lxmert}. These baselines and Switch-BERT follow the pre-train-then-fine-tune procedure with the controlled settings described above and have the same context for comparison.

Table \ref{tab:main_results} presents the experimental results of the model, together with results from these baselines. We observe that Switch-BERT has performances that are on par or better than the previous state-of-the-art architectures in these downstream tasks. The absolute improvements of 0.9\% on RefCOCO+, 1.8\% on VQAv2 and 1.1\% on Flick30K Image Retrieval over previous SOTA \footnote{For overall SOTA numbers that can be achieved without the controlled settings, readers can refer to \cite{yu2022coca} for VQAv2 and Flick30K Retrieval datasets, and \cite{kamath2021mdetr} for RefCOCO+.}  indicating that Switch-BERT can learn better vision and language representations that generalize better than these alternative methods to the downstream tasks. The controlled settings ensure the improvements are mainly contributed from the proposed architectures of Switch-Attention and Switch-Input blocks, which aim at easing the semantic misalignment between modalities and learning image-text modality interactions. 
Table \ref{tab:main_results} also includes the results of Switch-BERT without pre-training on Conceptual Captions dataset, i.e., initialized only from BERT in~\cite{devlin2018bert}. The degradation in performance demonstrates that the Switch-BERT benefits from pretraining as other multimodal BERTs.

\subsection{Ablation Studies}

\subsubsection{Effectiveness of the Switch-Attention and Switch-Input Blocks.} We start by investigating the influences of Switch-Attention and Switch-Input blocks. Following our controlled settings, we compare Switch-BERT with its three variants on downstream tasks. (i) SIB-ONLY: this variant uses normal encoder-style transformer layers instead of the Switch-BERT layer, (ii) SAB-ONLY: in this variant, we fix the input to each Switch-BERT layer to the output from its lower layer as usual. (iii) No-SIB-SAB: this variant is a normal single stream BERT encoder. All variants are evaluated following the pre-train-then-fine-tune procedure, and share the same hyperparameter setting with Switch-BERT.

Results in Table~\ref{tab:sab_sib_ablation} clearly show better performances by Switch-BERT than its variants. Given that SIB brings cross-layer input, we conclude that the semantic-level misalignment exists in single stream models and reducing misalignment between semantics of modalities results in better representations. The improvements of SAB-ONLY over the No-SIB-SAB variant also hint that our switching attention mechanism that learns to model modality associations is superior to the widely used single Joint-Attention mechanism. This bring us to the second question: Is the Joint-Attention necessary for Switch-Attention block?.

\setlength\tabcolsep{4pt}
\begin{table}[tp]
\setlength{\abovecaptionskip}{-0.5cm}
		\caption{An ablation study of interaction modes. Cross-Self (Self-Cross) and Joint stand for interaction modes. Pretraining indicates whether the models are pre-trained on Conceptual Captions before adapt to RefCOCO+. The default interaction mode is Self-Self Attention for all tested models, which means no interactions between modalities.} 
	\label{tab:joint_attention_ablation}
	\begin{center}
	\begin{tabular}{ccccc}
	\noalign{\smallskip}
	\hline
	\multicolumn{3}{c}{{Model}} & {RefCOCO+}  \\
	Pretraining & Cross-Self \& Self-Cross& Joint & Accuracy \\
	\noalign{\smallskip}
	\hline
	\checkmark&\checkmark&& 71.5  \\
	\checkmark& & \checkmark & 71.2  \\
	\checkmark&\checkmark & \checkmark & \textbf{72.8}  \\
	\hline
	&\checkmark&& 68.3  \\
	& & \checkmark & 67.5  \\
	&\checkmark & \checkmark & \textbf{68.9}  \\
	\hline
	\end{tabular}
	\end{center}
\end{table}

\begin{minipage}[htbp]{\textwidth}

\begin{minipage}[t]{0.48\textwidth}
\renewcommand\arraystretch{1.9}
\makeatletter\def\@captype{table}
\caption{An ablation study of Switch-Attention and Switch-Input blocks. } 
\label{tab:sab_sib_ablation}
\scalebox{0.5}{
\setlength{\tabcolsep}{3mm}{
\begin{tabular}{cccccc}
	\hline
	\multicolumn{2}{c}{{Model}} & \multicolumn{2}{c}{Flick30K} & VQAv2 & RefCOCO+ \\
	SIB& SAB& IR(r@1) & TR(r@1) &  VQA-score & Accuracy \\
	\hline
	&& 60.7 & 76.2 & 67.8 & 69.5  \\
	&\checkmark& 61.7 & 76.9 & 68.9 & 72.4 \\
	\checkmark& & 60.8 & 77.7 & 68.5 & 71.7   \\
	\checkmark&\checkmark & \textbf{62.2} & \textbf{78.2} & \textbf{70.7} & \textbf{72.8}  \\ 
	\hline
	\end{tabular}
 }
 }
\end{minipage}
\begin{minipage}[t]{0.48\textwidth}
\renewcommand\arraystretch{1.5}
\makeatletter\def\@captype{table}
\caption{An ablation study on effect of models' depth.} 
\label{tab:layer_ablation}
\scalebox{0.7}{
\begin{tabular}{cccc}
	\noalign{\smallskip}
	\hline
	\multicolumn{2}{c}{{Model}} & 
	VQAv2 &
	RefCOCO+ \\
	\multicolumn{2}{c}{\#layers} & 6 $\to$ 12  & 6 $\to$ 12  \\
	\noalign{\smallskip}
	\hline
	\multicolumn{2}{c}{UNITER}& 64.2 $\to$ 68.8 & 69.7 $\to$ 71.9  \\
	\multicolumn{2}{c}{Switch-BERT}& \textbf{65.4} $\to$ \textbf{70.7} & \textbf{70.2} $\to$ \textbf{72.8}   \\
	\multicolumn{2}{c}{SAB-ONLY}& 65.0 $\to$ 68.9 & 69.4 $\to$ 72.4  \\
	\hline
	\end{tabular}
 }
\end{minipage}
\end{minipage}

\subsubsection{Necessity of Joint-Attention.} 
We perform experiments on the RefCOCO+ dataset to verify the necessity for Joint-Attention. Table~\ref{tab:joint_attention_ablation} shows the results of Switch-BERT and its variants of the attention mode space with different initialization in the upper and lower panel. Models with the Cross-Self and Self-Cross Attention show similar results (71.5 vs 71.2) to those with Joint-Attention when pre-trained. However, even with Cross-Self \& Self-Cross, using Joint-Attention with negligible additional parameters consistently outperforms those without using it. 
Therefore, results support the necessity of Joint-Attention.

\subsubsection{Effect of Model's Depth.}
We also compare transferred results from models of varying depths including Switch-BERT and UNITER. Since Switch-BERT's SIB block introduces cross-layer connections given to more sensitivity to the model's depth, we also add the SAB-ONLY variant to the comparison. As shown in Table~\ref{tab:layer_ablation}, Switch-BERT of various depth show superior performance compared to its counterparts UNITER baseline. In addition, we observe meaningful improvements of Switch-BERT on the SAB-ONLY variant of fewer layers across multiple tasks evaluated, proving SIB help adapt to different tasks regardless of the model's depth. 

\subsubsection{Impact on scale of pre-training sets.}

\begin{figure}[htbp]
\setlength{\abovecaptionskip}{-0.3cm}
\begin{center}
   \includegraphics[width=1.0\linewidth]{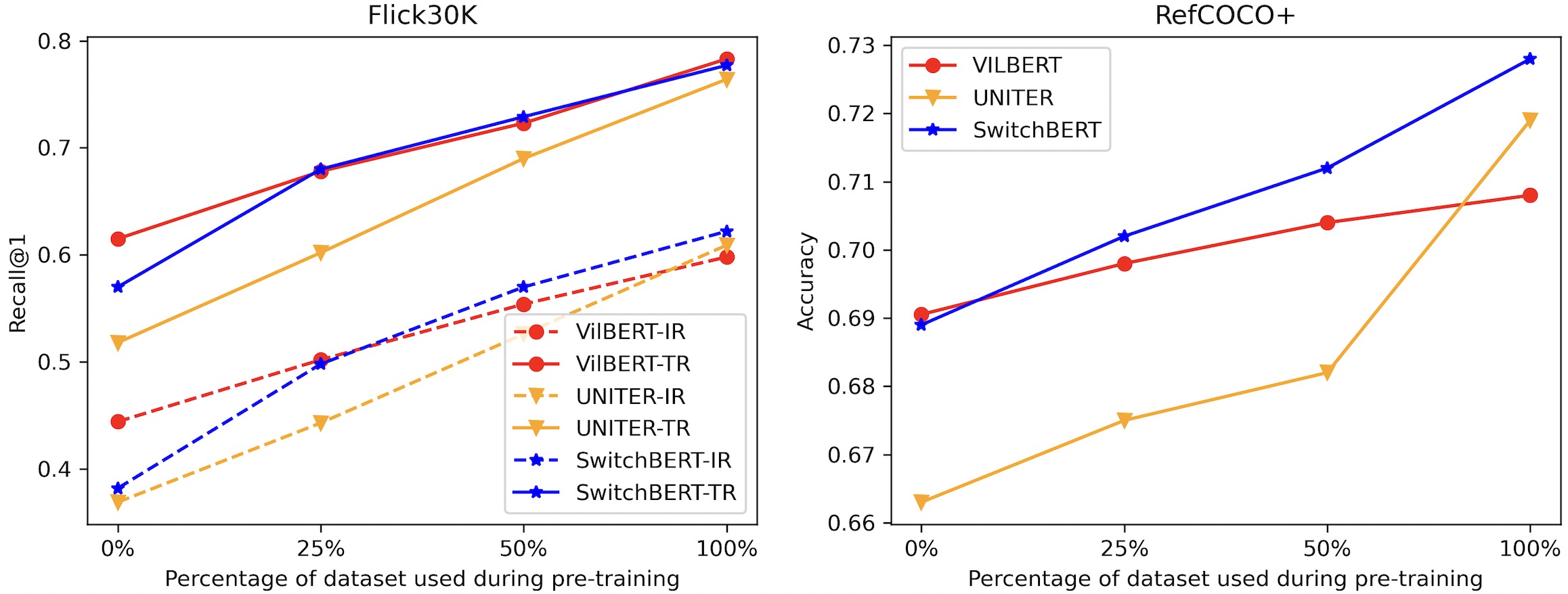}
\end{center}
   \caption{Effects on scale of pre-training sets. *-IR and *-TR represents the image-to-text retrieval and text-to-image retrieval tasks, respectively. We find large performance drop with less pre-training data for UNITER – implying single-stream models with only Joint-Attention ``eagers" for larger pre-training data before fully-trained.}
\label{fig:ablation_on_training_sets}
\end{figure}

\begin{table}[h]
\setlength{\abovecaptionskip}{-0.4cm}
	\caption{Computation overhead(FLOPs) and performances. Top-K routes are activated in Switch-Attention Blocks during fine-tuning. }
	\label{tab:training_cost}
	\begin{center}
	\begin{tabular}{cccccc}
	\noalign{\smallskip}
	\hline
	\textbf{Models} & 
	\multicolumn{2}{c}{VQAv2} & \multicolumn{2}{c}{RefCOCO+}\\
	\noalign{\smallskip}
        & FLOPs & VQA-score & FLOPs & Accuracy \\
	\hline
     UNITER
	& $2.31*1e^{16}$ & 68.8 & $3.68*1e^{15}$ & 71.9\\
	VilBERT
	& $2.72*1e^{16}$ & 68.7 & $4.29*1e^{15}$ & 70.8\\
	\hline
	Switch-BERT(K=4) &  $8.02*1e^{16}$ & 70.7 & $10.57*1e^{15}$ & 72.8  \\
        Switch-BERT(K=2) &  $3.07*1e^{16}$ & 70.2 & $5.27*1e^{15}$ & 72.1  \\
        Switch-BERT(K=1) &  $1.97*1e^{16}$ & 68.2 & $3.12*1e^{15}$ & 70.8  \\
	\hline
	\end{tabular}
    \end{center}
	\end{table}

We now turn our attention to the effect of pre-training dataset's scale on Switch-BERT's performance. For this experiment, we take random subsets of 25\% and 50\% from our conceptual caption dataset to pre-train models and then adapt them to various downstream tasks under our predefined controlled settings.  Shown in Fig.~\ref{fig:ablation_on_training_sets}, we can see that Switch-BERT benefits from increasing amounts of data as well as UNITER and VilBERT. Another observation is that larger performance gaps emerge between UNITER and VilBERT with less pre-training data on both evaluated tasks, we conjecture that UNITER(single-stream models) with only Joint-Attention eagers for larger pre-training data volumes to get fully-trained, and Switch-BERT alleviates this problem with complete interaction mode space.

\subsubsection{Computation overhead of Switch-BERT.}
We estimate the number of floating point operations of training a model on each downstream task for static approaches. For Switch-BERT, we track its routing path and accumulate the operation count during training due to its dynamic characteristics. Results are shown in Table~\ref{tab:training_cost}.  Switch-BERT indeed requires extra computation to converge compared to traditional static models, the overhead is mainly caused by SABs that activate all paths at the beginning of training. We further investigate this with only top-K paths in SAB activated, and observe an acceptable overhead and performance balance when K=2.

\subsection{Qualitative Studies.}

\begin{figure*}[tbp]
\setlength{\abovecaptionskip}{-0.5cm}
\begin{center}
\includegraphics[width=0.8\linewidth,height=0.3\linewidth]{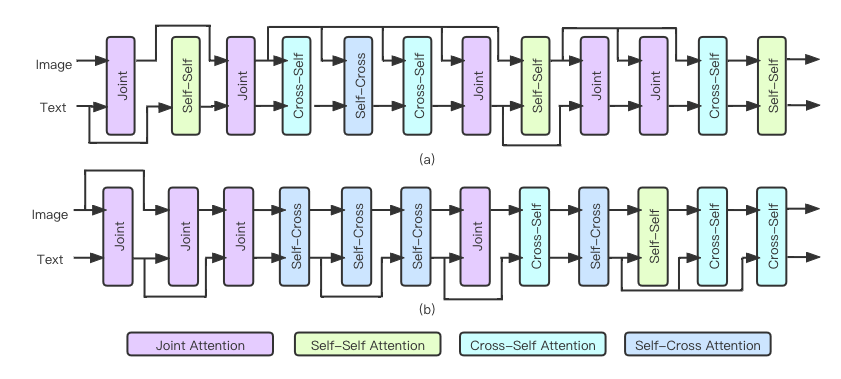}
\end{center}
\caption{Architectures Learned by Switch-BERT. (a) and (b) shows the learned architectures for referring expression comprehension and cross-modal retrieval tasks, respectively.}
\label{fig:learned_architecture}
\end{figure*}

With SAB and SIB modules, Switch-BERT should be able to adapt its architecture to different multimodal tasks. To confirm this, we analyze utility of SAB and SIB on Switch-BERT fine-tuned on referring expression comprehension and cross-modal retrieval tasks. We sort the learned architectures according to their occurrence frequency on each task.   
Fig.~\ref{fig:learned_architecture} illustrates the most frequently used architectures by Switch-BERT on the two tasks. For the referring expression comprehension task,  Switch-BERT learns to use cross-layer representations more frequently for the visual modality than the text modality. 
On the cross-modal retrieval task, Switch-BERT uses Self-Self attention once but more frequently with other attention modes that involve interactions between modalities. 
The frequencies of selecting these most-frequent architectures are dominantly 48.79\% and 31.09\%, respectively, on the two tasks. These results indicate that Switch-BERT is able to extract task-specific architecture.

\section{Related Work}
\subsubsection{Multimodal BERTs.} BERT-style representations \cite{devlin2018bert,liu2019roberta,yang2019xlnet,lan2019albert,dong2019unified} have been advancing the state-of-the-art performances in natural language processing in recent years. Its success has encouraged researchers to apply them  more widely to tasks including multimodality. Methods based on BERT architectures have been proposed recently and have become the dominant approaches in applications such as video captioning~\cite{sun2019videobert}. The works of VisualBERT~\cite{li2019visualbert}, UNITER~\cite{chen2020uniter}, VLBERT~\cite{su2019vl}, and PixelBERT~\cite{huang2020pixel} employ a single-stream BERT encoder for joint modeling of interactions between modalities. The other dual-stream approach including ViLBERT~\cite{lu2019vilbert} and LXMERT~\cite{tan2019lxmert} has representations separately for each modality and uses cross-attention mechanism to model their interactions. The proposed method distinguishes from the above methods in using flexible Switch Attention-and-Input mechanism to select proper interaction modes and cross-layer input. It aims at alleviating the not-well-studied semantic misalignment problem. Empirically, we have confirmed its superior performances over the other methods.

\subsubsection{Conditional Computation Models.} The proposed Switch-BERT dynamically adjusts its architecture according to inputs. It is therefore in line with Mixture of Experts (MoE) methods in~\cite{shazeer2017outrageously,fedus2021switch}. The method in \cite{shazeer2017outrageously} uses a gating function to select experts to perform computations. The method in \cite{fedus2021switch} introduces the MoE layer into the Transformer architecture and applies a routing algorithm that sends tokens to their token-specific experts. Switch-BERT differs from these works in two aspects: i) instead of using MoE as a substitute of the FFN layers in~\cite{shazeer2017outrageously}, it selects sample-specific attention and input with the novel Switch Attention-and-Input blocks;
ii) whereas MoE is conducted at token-level in \cite{fedus2021switch}, Switch-BERT conducts switch operations at modality-level and cross-layer. Besides, our switch input mechanism learns to “select" or “skip" a transformer layer, which shares the same spirit with variable depth in Transformer\cite{li2020deep}. Work in \cite{li2020deep} explores using a shared deep Transformer for multiple tasks with the learned distribution of layer selection, the learned distribution is restricted on task-level. While for Switch-BERT, the layer selection distribution is conditioned on modality inputs, such that it performs sample-specific switch operations. To our best knowledge, Switch-BERT is the first attempt to have conditional computation for multimodal learning.

\section{Conclusion}
In this paper, we proposed Switch-BERT to effectively alleviate the modality misalignment problem for multimodal representation learning. Switch-BERT learns to model intra- and inter-modal interactions and select interaction mode for each layer individually. It also learns to select, for each layer, the inputs that are not restricted to the current layer and therefore learns selecting inputs cross layers.  We verified its effectiveness through controlled settings on multimodal tasks including visual question answering, cross-modal retrieval, and referring expression comprehension. We also carried out ablation studies to confirm that Switch-BERT is capable of learning task-specific architectures. Experimental results show that Switch-BERT dynamically adapts its structure and consistently achieve better or comparable performances than other state-of-the-art fixed architectures on a variety of multimodal tasks. In future work, we plan to explore the efficiency of variant mechanisms and reveal the internal alignment with more details.

\noindent\textbf{Acknowledgments.} The authors would like to thank the anonymous reviewers for their helpful feedback that improved this work.

\clearpage
%
%
\bibliographystyle{splncs04}
\bibliography{egbib}
\end{document}